\documentclass[letterpaper, 10 pt, journal, twoside]{IEEEtran}
\usepackage{amssymb}
\usepackage{bbding}
\usepackage{multirow}
\usepackage{xcolor}
\usepackage{bm}
\definecolor{darkred}{rgb}{0.8, 0.0, 0.0}

\usepackage[pdftex]{graphicx}
\usepackage{cite}

\usepackage{etoolbox}
\AtBeginEnvironment{figure}{ % 
  \setlength{\abovecaptionskip}{-6pt} % 
}

\AtBeginEnvironment{table}{ % 
  \setlength{\abovecaptionskip}{-2pt} % 
}
\AtBeginEnvironment{table*}{ % 
  \setlength{\abovecaptionskip}{-2pt} % 
}
\AtBeginEnvironment{equation}{
\setlength{\abovedisplayskip}{2pt}
\setlength{\belowdisplayskip}{2pt}
}

\ifCLASSINFOpdf
\else
\fi
\usepackage{amsmath}
\begin{document}
\title{Improving Hierarchical Representations of Vectorized HD Maps with Perspective Clues}

% \author{Michael~Shell,~\IEEEmembership{Member,~IEEE,}
%         John~Doe,~\IEEEmembership{Fellow,~OSA,}
%         and~Jane~Doe,~\IEEEmembership{Life~Fellow,~IEEE}% <-this % stops a space
% \thanks{M. Shell was with the Department
% of Electrical and Computer Engineering, Georgia Institute of Technology, Atlanta,
% GA, 30332 USA e-mail: (see http://www.michaelshell.org/contact.html).}% <-this % stops a space
% \thanks{J. Doe and J. Doe are with Anonymous University.}% <-this % stops a space
% \thanks{Manuscript received April 19, 2005; revised August 26, 2015.}}
\author{Chi Zhang$^{1}$, Qi Song$^{1}$, Feifei Li$^{1}$, Jie Li$^{2}$, and Rui Huang$^{1}$
% First A. Author$^{1}$, Second B. Author$^{2}$, and Third C. Author$^{1}$%
\thanks{Manuscript received: June 18, 2025; Revised September 3, 2025; Accepted September 29, 2025. This paper was recommended for publication by Editor Javier Civera upon evaluation of the Associate Editor and Reviewers' comments.
This work supported in part by Shenzhen Science and Technology Program under Grant JCYJ20220818103006012,  KJZD20240903100202004, and 20231128093642002, and Guangdong Basic and Applied Basic Research Foundation under Grant 2023A1515110729. \textit{(Corresponding authors: Rui Huang; Jie Li.)}} %Use only for final RAL version
\thanks{$^{1} $Chi Zhang, Qi Song, Feifei Li, and Rui Huang are with School of Science and Engineering, The Chinese University of Hong Kong (Shenzhen), Longgang, Shenzhen, Guangdong, 518172, P.R. China (e-mail: chizhang1,feifeili1,qisong@link.cuhk.edu.cn; ruihuang@cuhk.edu.cn)}

\thanks{$^{2} $Jie Li is with the Undergraduate School of Artificial Intelligence, Shenzhen Polytechnic University, Shenzhen, Guangdong, 518055, P.R. China (e-mail: jieli1@szpu.edu.cn)}%
\thanks{Digital Object Identifier (DOI): see top of this page.}
}
\maketitle

 \begin{figure*}[t]

    \centering
    \includegraphics[width=0.92\linewidth]{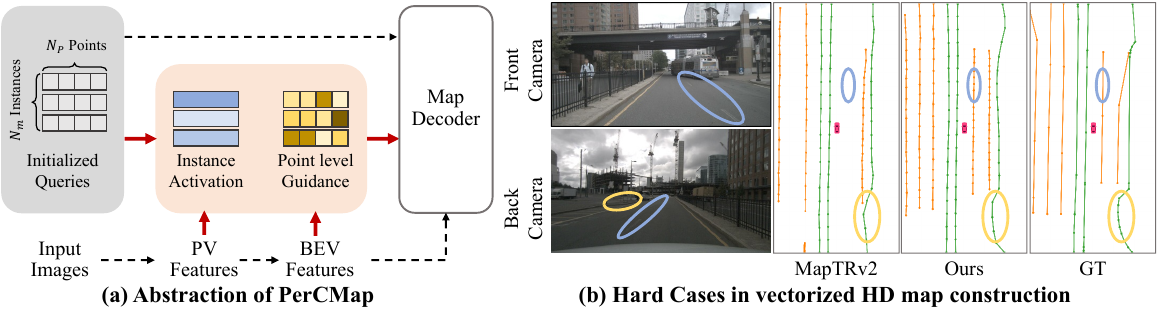} \vspace{-10pt}
    \caption{\textbf{Abstract pipeline and prediction samples of PerCMap.} In (a), \(\dashrightarrow\) indicates the flow of the traditional sequential pipeline, whereas \textcolor{darkred}{\(\bm{\rightarrow}\)} indicates our redesigned flow that reuses PV features at both instance and point levels. In (b), MapTRv2\protect\cite{liao2023maptrv2} fails to preserve instance clues and shape priors in the circled regions, while PerCMap accurately detects both lane markings and boundary shapes. } % \vspace{-18pt}

    \label{fig:intro}
\end{figure*}

\begin{abstract}
The construction of vectorized High-Definition (HD) maps from onboard surround-view cameras has become a significant focus in autonomous driving. However, current map vector estimation pipelines face two key limitations: input-agnostic queries struggle to capture complex map structures, and the view transformation leads to information loss. These issues often result in inaccurate shape restoration or missing instances in map predictions. To address this concern, we propose a novel approach, namely \textbf{PerCMap}, which explicitly exploits clues from perspective-view features at both instance and point level. Specifically, at instance level, we propose Cross-view Instance Activation (CIA) to activate instance queries across surround-view images, thereby helping the model recover the instance attributes of map vectors. At point level, we design Dual-view Point Embedding (DPE), which fuses features from both views to generate input-aware positional embeddings and improve the accuracy of point coordinate estimation. Extensive experiments on \textit{nuScenes} and \textit{Argoverse 2} demonstrate that PerCMap achieves strong and consistent performance across benchmarks, reaching 67.1 and 70.5 mAP, respectively.
\end{abstract}

% \begin{IEEEkeywords}
% IEEE, IEEEtran, journal, \LaTeX, paper, template.
% \end{IEEEkeywords}

% \fi
\IEEEpeerreviewmaketitle

\section{Introduction}
% for IEEE journal papers produced under \LaTeX\ using
% IEEEtran.cls version 1.8b and later.
\IEEEPARstart{M}{ap} construction in Bird’s-Eye-View (BEV) using onboard sensors plays a crucial role in autonomous driving. Existing methods typically employ either pixel-wise semantic maps \cite{lu_monocular_2019,roddick_predicting_2020,yang2021pyva,zhou2022cross} or vectorized High-Definition (HD) maps \cite{li2022hdmapnet,liu2023vectormapnet,liao2022maptr,ding2023pivotnet,liao2023maptrv2,map:zhou2024himap,choi2024mask2map}. Compared to semantic maps, vectorized HD maps encode map elements as point sets, offering greater sparsity and improved scalability. We focus on constructing such vectorized HD maps from surround images.

In vectorized maps, each map vector is represented as a combination of a categorical label and an ordered point set, essentially forming a classified polygon or polyline. Existing methods typically model this representation by decoupling the instance level semantics and coarse position from the fine-grained point level geometry, and then use transformer-based networks for map prediction \cite{liu2023vectormapnet,liao2022maptr}. For example, VectorMapNet \cite{liu2023vectormapnet} employs a two-stage architecture that first localizes instances coarsely, then predicts precise point coordinates. More typically, MapTR \cite{liao2022maptr} and many follow-up works \cite{liao2023maptrv2,yang2024mgmapnet,map:zhou2024himap} utilize a hierarchical querying mechanism with dedicated instance and point queries (as shown in the gray area in Fig. \ref{fig:intro}(a)). Meanwhile, these methods adopt a sequential-form pipeline (indicated by dashed arrows in Fig. \ref{fig:intro}(a)), which first converts image features from Perspective-View (PV) to BEV, and then decodes the map using only BEV features. Nevertheless, this pipeline has two main limitations. First, the input-agnostic hierarchical queries constrain the model’s learning capability in complex scenarios. These queries must infer precise geometric details without leveraging any input priors, which often leads to failure in capturing fine-grained map structures. As highlighted by the yellow circle of Fig. \ref{fig:intro}(b), MapTRv2 incorrectly predicts the shape of a road bulbout. Second, there exist unavoidable information losses in view transformation from PV to BEV. Since decoding relies solely on BEV features, this loss remains unresolved and can ultimately cause misalignment with the original inputs. This is evident in the blue ellipse of Fig. \ref{fig:intro}(b), where prior work fails to detect map elements that are visible in the raw inputs.

An intuitive solution is to directly reuse PV features on the decoder queries. For instance, recent studies have demonstrated the effectiveness of encoding PV proposals or semantic predictions into object queries to enhance 3D object detection \cite{yang2023bevformerv2,song2024sdtr}. However, this strategy is less effective for vectorized map construction. Since map elements are strictly ground-level, PV features often contain redundant regions, such as the sky or buildings. Moreover, map vectors involve hierarchical representations, including both instance semantics and point level geometry, which require tailored designs for effective PV feature reuse. These challenges highlight the need for customized perspective-view information exploitation to support vector map representation.

In this paper, we propose a novel approach for HD Map construction by perspective clue re-utilization, termed \textbf{PerCMap}, as abstracted in Fig. \ref{fig:intro}(a). PerCMap leverages PV information at both the instance and point levels. At instance level, we introduce Cross-view Instance Activation (CIA), which produces robust instance-aware queries via aggregating surround-view PV features. This facilitates the recognition of complete map instances that may span multiple camera views. At point level, we design Dual-view Point Embedding (DPE), which captures point level decoding guidance by modeling interactions between PV and BEV features. DPE generates input-dependent positional embeddings while simultaneously enhancing the expressiveness of BEV features. The proposed PerCMap re-leverages PV at both instance and point levels, effectively improves the map estimation accuracy. As shown in Fig. \ref{fig:intro}(a), PerCMap demonstrates its ability to preserve the initial priors from the original images, surpassing the performance of previous method like MapTRv2. 

To summarize, the contributions of this work are as follows:
\begin{itemize}
    \item We propose a novel vectorized map construction pipeline by reintroducing perspective-view information to enhance both instance and point level vector representation.
    \item We design two complementary modules: Cross-view Instance Activation (CIA) aggregates surround-view features to generate instance-aware queries, while Dual-view Point Embedding (DPE) enriches point level decoding with input-conditioned spatial cues.
    \item Experimental results on standard benchmarks show that the proposed method outperforms existing approaches, validating its effectiveness.
\end{itemize}


\section{Related Works}
\subsection{Vectorized HD Map Construction}

Recent works on HD map construction increasingly adopt a vectorized representation, where map elements such as lane dividers or boundaries are modeled as polylines or polygons composed of ordered points and categorical labels \cite{liu2023vectormapnet,liao2022maptr,liao2023maptrv2,ding2023pivotnet,liu2024mgmap,liu2024mapqr,qiu2024ticmapnet,choi2024mask2map}. This structure reflects a hierarchical nature: instance level semantics define object type or coarse layout, while point level features capture precise geometry necessary for shape reconstruction. Estimating map vectors is challenging as it requires jointly modeling categorical attributes and fine-grained spatial structures. To handle multi-camera input, most existing methods \cite{li2022hdmapnet,liu2023vectormapnet,liu2024mapqr} adopt a sequential pipeline that transforms PV features into the BEV space, followed by vectorized decoding. While this design enables unified spatial reasoning, it often causes information loss. Visual cues specific to perspective view, such as lane continuity and boundary shapes, may be degraded or discarded during projection. These two factors, the structural complexity of vectorized maps and the limitations of view transformation, affect the quality of final predictions. Several recent studies attempt to address these challenges from different angles. MGMapNet \cite{liu2024mgmap} models map vectors by aggregating coarse instance level and fine-grained point level queries from multi-scale BEV features. HIMap \cite{map:zhou2024himap} designs an interactor to explicitly extract features regarding point position and element shape. However, both methods operate entirely within the BEV space and do not explicitly recover the visual information lost in the PV-to-BEV transformation.

In this work, PerCMap retains the instance-point query paradigm and enhances both instance and point level prediction by reusing PV features. This perspective-aware feature extraction complements BEV features and improves the overall accuracy of map construction.

%\IEEEpubidadjcol

\subsection{BEV Scene Parsing}
Predicting scene content in the BEV from surround-view PV images is central to autonomous driving perception. Early methods like DETR3D \cite{wang2022detr3d} sample PV image features using 3D spatial relationships to produce BEV predictions. More recent approaches apply PV-to-BEV transformations to synthesize BEV features, as in LSS \cite{philion2020lift}, which estimates latent depth to compensate for missing 3D cues. Transformer-based models further enhance this process through query generation and attention mechanisms \cite{li2022bevformer,liu2022petr}.

This transformation pipeline has been widely adopted in BEV-based tasks such as object detection \cite{liu2022petr,yang2023bevformerv2,song2024sdtr}, segmentation \cite{zou2023hft,roddick_predicting_2020}, and map construction \cite{li2022hdmapnet,li2023bimapper,liu2024mapqr}. Several works have addressed information loss in view transformation by re-leveraging PV features. BEVFormerV2~\cite{yang2023bevformerv2} and SDTR~\cite{song2024sdtr} predict bounding box or semantic logits in PV to provide priors for the decoding of 3D objects, while DuoSpaceNet~\cite{huang2025duospacenet} applies deformable attention jointly on PV and BEV duo space. In temporal multi-frame perception, SOLOFusion~\cite{park2023solofusion} builds a multi-frame PV cost volume to predict depth for BEV construction, and TempBEV~\cite{monninger2024tempbev} aggregates multi-frame PV features with cross attention to update BEV representations. Beyond BEV-based methods, Anchor3DLane++~\cite{huang2024anchor3dlane++} directly samples PV features with 3D anchors to estimate lanes without BEV lifting. Despite these advances, PV-based BEV enhancement has yet to be explored for HD map construction, where directly importing PV proposals or features may also introduce redundancy.

In HD map construction, which targets static ground-level elements, PV-to-BEV transformation is typically applied after the image backbone to extract BEV features. While general-purpose view transformations \cite{philion2020lift,li2022bevformer} are often used, recent works have also proposed map-specific variants \cite{qiao2023bemapnet,liu2024mapqr}. Yet most pipelines rely solely on BEV features for decoding, which can lead to degraded predictions due to early-stage information loss. To address this, we propose re-utilizing PV features to directly support the decoding of map vectors.

\begin{figure*}[t]
    \centering
    \includegraphics[width=0.95\linewidth]{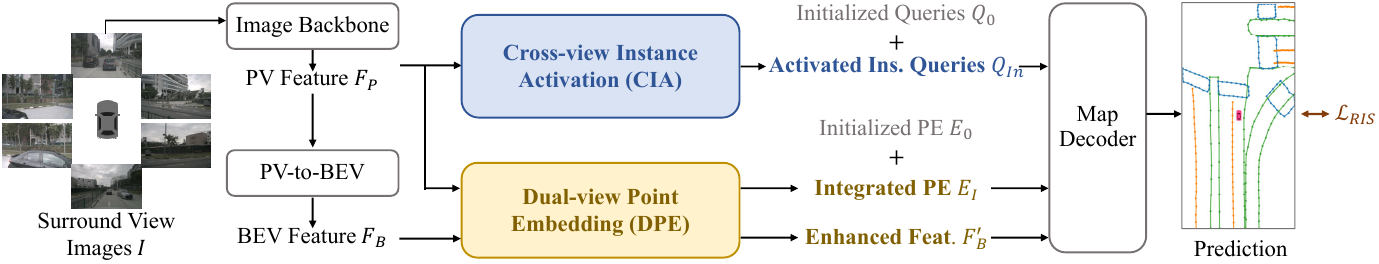} \vspace{-10pt}
\caption{\textbf{Overall architecture of PerCMap.} The proposed Cross-view Instance Activation (CIA) and Dual-view Point Embedding (DPE) enhance vector prediction at the instance and point levels, respectively. Unlike the typical sequential pipeline, PerCMap effectively reuses PV features to improve prediction accuracy and map construction quality. ``PE" refers to positional embedding. } % \vspace{-8pt}
    \label{fig:overall}
\end{figure*}

\section{Method}

\subsection{Overview}
The pipeline of PerCMap is illustrated in Fig. \ref{fig:overall}. Given RGB images \(\{I_i \in \mathbb{R}^{H_I \times W_I \times 3}\}_{i=1}^N\) from \(N\) cameras, the model predicts \(N_m\) vectorized map instances, each with \(N_p\) points \(\{\hat{v}_i \in \mathbb{R}^{N_p \times 2}\}_{i=1}^{N_m}\) and class scores \(\{\hat{c}_i \in \mathbb{R}^{N_c \times 1}\}_{i=1}^{N_m}\) over \(N_c\) categories.

\textbf{Image backbone and PV-to-BEV. }
The input images are sent into the image backbone and converted to PV features \(F_P\in \mathbb{R}^{N\times H_p\times W_p\times C}\). The PV-to-BEV module further transforms the PV features into BEV features \(F_B\in \mathbb{R}^{H\times W\times C}\).  

\textbf{Cross-view Instance Activation (CIA). }
CIA encodes new instance queries \(Q_{In} \in \mathbb{R}^{N_m \times C}\) from surround-view PV features \(F_P\). Details are provided in Sec.~\ref{sec:cvia}.

\textbf{Dual-view Point Embedding (DPE). }
DPE fuses \(F_P\) and \(F_B\) to generate an integrated heatmap, from which it derives positional embeddings \(E_I\), and enhances BEV features to \(F_B'\). Further explanation is in Sec.~\ref{sec:dem}.

\textbf{Map Decoder. }
Features from CIA and DPE are passed to the map decoder. The Activated Instance Query \(Q_{In}\) and Integrated Positional Embedding \(E_I\) are combined with the initialized query \(Q_0\) and positional embedding \(E_0\), respectively. In addition, the enhanced BEV feature \(F_B'\) from DPE is used as the decoder’s value input. The decoder outputs \(N_m\) map elements, each with predicted coordinates \(\hat{v}\) and classification scores \(\hat{c}\).

\textbf{Losses.} We apply a heatmap loss \(\mathcal{L}_H\) to guide DPE and a rasterized instance segmentation loss \(\mathcal{L}_{RIS}\) to supervise final predictions. See Sec.~\ref{sec:loss} for details.

We introduce CIA, DPE, and \(\mathcal{L}_{RIS}\) to enhane the performance of map construction. CIA and DPE establish direct connections between PV features and the map decoder. This design mitigates error accumulation from the PV-to-BEV transformation and enables the targeted extraction of instance level and point level map features.

\subsection{Cross-view Instance Activation} \label{sec:cvia}

\begin{figure}[t]
% \vspace{-7pt}
    \centering
    \includegraphics[width=0.95\linewidth]{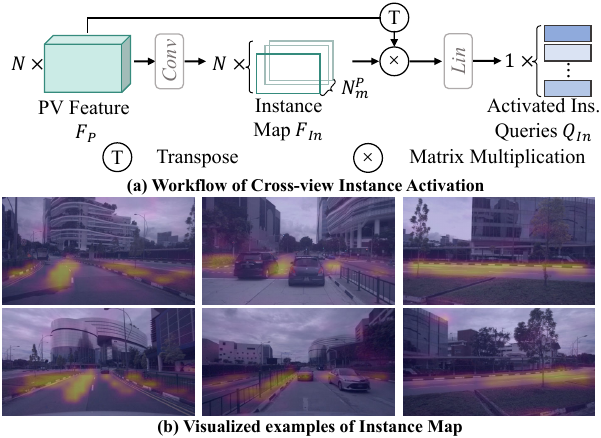}
    % \vspace{-20pt}
     \caption{\textbf{Process and visualized examples of Cross-view Instance Activation.} (a) CIA extracts and aggregates instance features from multiple views to generate the Activated Instance Query. (b) Heatmap of the Instance Map \(F_{In}\), where high intensities align with map element regions.}    % \vspace{-16pt}
    \label{fig:ins}
\end{figure} 

Inspired by the Instance Activation Map (IAM) \cite{cheng2022sparse}, we aim to highlight instance regions and attributes based on PV image features. However, directly applying IAM is unsuitable for the HD map construction tasks, which involve surround-view feature aggregation and view transformation. To address this, we refine and integrate surround-view PV features to initiate instance activation, creating the Activated Instance Query \(Q_{In}\) in a cross-view manner. We refer to this process as Cross-view Instance Activation (CIA). 

As illustrated in Fig.~\ref{fig:ins} (a), CIA first generates per-view instance maps from PV features:
\begin{equation}
    F_{In} = \sigma(\text{Conv}(F_P))
\end{equation}
where \(F_{In} \in \mathbb{R}^{N \times N^P_m \times C}\), and \(\sigma\) denotes sigmoid activation. \(N^P_m\) is the predefined number of activated instances per view.

To transform per-view instance cues into BEV-based instance queries, we compute
\begin{equation}
    Q_{In} = Lin(F_{In} \times F_P^{\intercal})
\end{equation}
where \(Lin(\cdot)\) is a linear layer. The final instance queries used in decoding is:
\begin{equation}
    Q = Q_0 + Q_{In}
\end{equation}
where \(Q_0\) is the initialized queries.

Unlike prior IAM-based approaches in driving perception \cite{luo2023latr,liu2024mgmap}, CIA requires no extra supervision or instance matching, yet still enhances instance-aware representation. Fig. \ref{fig:ins}(b) visualizes the examples of Instance Map \(F_{In}\), where features near map element regions are highlighted, increasing their likelihood of being transferred in the next step to form the final instance queries. 

% \vspace{-8pt}
\subsection{Dual-view Point Embedding} 
\label{sec:dem}
\begin{figure}[t]
    \centering
    \includegraphics[width=0.95\linewidth]{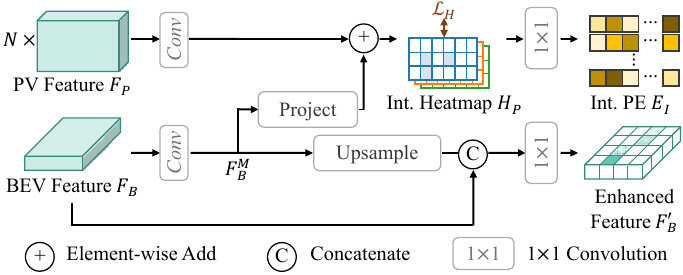}% \vspace{-12pt}
     \caption{\textbf{Workflow of Dual-view Point Embedding.}  Given the features of both views, DPE explicitly incorporates these, with the intermediate Integrated Heatmap is supervised. BEV features are thereby enhanced due to the influence of projection and supervision.} % \vspace{-12pt}
    \label{fig:de}
\end{figure}

The Dual-view Point Embedding (DPE), shown in Fig. \ref{fig:de}, enhances point level decoding by integrating priors from both PV and BEV features to generate input-dependent positional embeddings. An intermediate output, the Integrated Heatmap, is supervised to further improve BEV feature representation and decoding accuracy.

\subsubsection{Integrated Positional Embedding}
In this branch of DPE, BEV features are projected to the perspective view and fused with PV features to generate Integrated Positional Embeddings. Following transformer-based decoder designs\cite{det:zhu2020deformable}, these embeddings encode spatial relationships and serve as point level guidance for sampling and attention mechanisms. Unlike conventional input-agnostic embeddings, our design incorporates dual-view cues, i.e. PV and BEV, to provide spatial priors for the map decoder, improving point level decoding capability.

We first downsample \(F_B\) into the BEV Enhancement Map \(F_B^{M}\) from BEV features by multi-layer convolution:
\begin{equation}
\label{eq:emap}
    F_B^{M} = Conv(F_B)
\end{equation}
where \(F_B^{M} \in \mathbb{R}^{N\times(H/d)\times (W/d)\times C}\), \(d\) is the predefined downsample scale factor. \(F_B^{M}\)  is ready for the projection into PV.

For an arbitrary coordinate \((x,y)\) in the BEV plane, we first complete it to a 3D coordinates \((x,y,z)\) by assigning the height value. Given the camera parameter matrix \(\boldsymbol{P}\),  the coordinate \((x,y,z)\) in the 3D camera ego space can be converted to pixel \((u,v)\) in the image coordinate as follows: 
\begin{equation}
\label{eq:proj}
    [u,v,1]^{T} = \boldsymbol{P}\times[x,y,z,1]^T
\end{equation}
We then create a set perspective-view canvas for each class, denoted as \(F_P^{M}\in \mathbb{R}^{N\times H_I\times W_I \times N_c}\) and initialize them as blank. \(F_B^{M}\), generated from Eq.\ref{eq:emap}, is firstly convoluted into category-wise features, and then projected by assigning feature values to the corresponding pixel in the canvas: 
\begin{equation}
    F_P^{M}[u,v,c^{id}]  = Conv(F_B^{M})[x,y,c^{id}]
\end{equation}
in which the relationship between \((u,v)\) and \((x,y)\) is illustrated in Eq. \ref{eq:proj}, and \(c^{id}\) is the category index. The Integrated Heatmap \(\hat{H_P}\) is obtained by aggregating the PV features and the projected perspective-view feature canvas:
\begin{equation}
    \hat{H_P} = Conv(F_P)+Conv(F_P^M)
\end{equation}
Ultimately, the Integrated Positional Embedding \(E_I\) is encoded from the Integrated Heatmap via convolution of kernel size \(1\times 1\). The final positional embedding is the sum of \(E_I\) and initialized positional embeddings \(E_0\):
\begin{equation}
    E =E_0 +E_I=E_0+Conv_{1\times 1}(H^P)
\end{equation}

Integrated Heatmap \(\hat{H_P}\) serves as an essential intermediate during the synthesis of  \(E_I\). To guide the positional embeddings to be aware of points features, we supervise \(\hat{H_P}\) by the visible \textit{keypoints} in the perspective-view and encapsulate the supervision as a part of DPE. To be more specific, \textit{keypoints} refer to the annotated points in the simplest representation of the map instances, defining the geometric shape of the map elements. The details will be introduced in Sec. \ref{sec:loss}.

\subsubsection{BEV Feature Enhancement}
Previously, we mentioned that the BEV Enhancement Map  \(F_B^{M}\) s derived from the BEV features, which are obtained through the view transformer. Additionally, \(F_B^{M}\) contributes to the formation of the Integrated Heatmap, which is under supervised. We believe that \(F_B^{M}\) integrates information strongly related to map elements within the BEV features and anticipate its continued utilization to enhance the representational effectiveness of these BEV features, thereby further optimizing subsequent decoding processes. 

Given the BEV Enhancement Map derived from Eq. \ref{eq:emap}, the enhancement on BEV features is exclusively carried out within the domain of BEV. The BEV features \(F^{\prime}_B\) are enhanced by collapsing the \(F_B\) and \(F_B^{M}\):
\begin{equation}
        F_B^\prime  = Conv_{1\times1}(Cat(F_B,\Omega(F_B^{M})))
\end{equation}
where ``\(\Omega(\cdot)\)" is the upsample function and ``\(Cat\)" is concatenation. The enhanced BEV features \(F^{\prime}_B \in \mathbb{R}^{H\times W\times C}\) eventually serve as the input value of the map element decoder. 

\subsection{Training Loss} \label{sec:loss}

\subsubsection{Heatmap Loss \(\mathcal{L}_{H}\)}
Given ground-truth keypoints projected onto PV images, the target heatmap \(H_P\) is generated using a Gaussian kernel, following keypoint detection methods~\cite{det:law2018cornernet,det:duan2019centernet}. The heatmap loss is computed as:
\begin{equation}
    \mathcal{L}_{H} = \mathcal{L}_{Focal}(\text{Sigmoid}(\hat{H}_P), H_P),
\end{equation}
where \(\mathcal{L}_{Focal}\) denotes focal loss~\cite{det:lin2017focal} between predicted and target heatmaps.

\subsubsection{Rasterized Instance Segmentation Loss \(\mathcal{L}_{RIS}\)}

To encourage the model to focus on restoring the geometric shape in map prediction, we introduce the Rasterized Instance Segmentation (RIS) loss  \(\mathcal{L}_{RIS}\).

For a predicted vector \(\hat{v}_i\) with class score \(\hat{c}_i\), we assign scores to corresponding pixels to form a BEV mask \(\hat{M}_R\). The binary target mask \(M_R\) is generated from ground-truth labels \(c_i\). The RIS loss is defined as:
\begin{equation}
    \mathcal{L}_{RIS} = \mathcal{L}_{CE}(\hat{M}_R, M_R),
\end{equation}
where \(\mathcal{L}_{CE}\) is the cross-entropy loss.

\subsubsection{Total Loss}

Following~\cite{liao2022maptr}, we apply focal loss and L1 loss to supervise category prediction and point locations, denoted as \(\mathcal{L}_{cls}\) and \(\mathcal{L}_{pts}\), respectively. Additionally, we adopt the auxiliary loss set \(\mathcal{L}_{BS}\) from~\cite{liao2023maptrv2}, including edge direction, one2many, and dense supervision. The total loss is:
\begin{equation}
    \mathcal{L}_{total} = \gamma_H \mathcal{L}_H + \gamma_R \mathcal{L}_{RIS} + \mathcal{L}_{cls} + \mathcal{L}_{pts} + \mathcal{L}_{BS},
\end{equation}
where \(\gamma_H\) and \(\gamma_R\) are weighting factors.

\begin{table*}[ht]
    \caption{Comparison with SOTA methods on \textit{nuScenes} val set. ``EB0" is the backbone of previous work \protect\cite{efficientnet2019}. ``\dag" indicates the results are re-evaluated by publicly available models. All FPS results are re-tested on the same device. The best and second best results are highlighted in \textbf{bold} and \underline{underline}, respectively.  ``*" indicates the same configuration used in ablation studies.}
    \centering
    \begin{tabular}{lccc|cccc|cccc|c}
        \hline
        % Method  & Backbone & Epochs & \(\mbox{AP}_{ped}\) & \(\mbox{AP}_{div}\) & \(\mbox{AP}_{bdr}\) &  $mAP$ & & & &\\
        \multirow{2}{*}{Method}  & \multirow{2}{*}{Bkb.}& \multirow{2}{*}{Epo.}  &\multirow{2}{*}{\(N_m\)}& \multicolumn{4}{c|}{Coarse Thresholds} & \multicolumn{4}{c|}{Tight Thresholds} & \multirow{2}{*}{FPS} \\ 
                & &  && \(\mbox{AP}_{ped}\) & \(\mbox{AP}_{div}\) & \(\mbox{AP}_{bdr}\) &  \(\mbox{mAP}\)& \(\mbox{AP}_{ped}\) & \(\mbox{AP}_{div}\) & \(\mbox{AP}_{bdr}\) &\(\mbox{mAP}_{T}\) & \\
        \hline
 HDMapNet\cite{li2022hdmapnet}    & EB0 &30  &-&14.4& 21.7& 33.0&23.0& 7.1& 28.3& 32.6& 22.7& -\\
 VectormapNet\cite{liu2023vectormapnet}     & R50  & 110  &35&36.1& 47.3& 39.3&40.9 & 18.2\dag& 27.2\dag& 18.4\dag& 21.3\dag &-\\

 MapTR\cite{liao2022maptr}& R50& 24 &50& 46.3& 51.5& 53.1&50.3& 23.2\dag& 30.7\dag& 28.2\dag&27.3\dag &14.5\\
 BeMapNet\cite{qiao2023bemapnet}& R50& 30  &60& 57.7 &62.3 &59.4 &59.8 &\underline{39.0}&\textbf{46.9}&37.8 &\underline{41.3}&-\\
 % BeMapNet~\shortcite{qiao2023bemapnet}& R50& 30&57.7& 62.3& 59.4&59.8& 39.0& \textbf{46.9}& 37.8& \textbf{41.3}\\
 PivotNet\cite{ding2023pivotnet}& R50& 24  &60& 56.2& 56.5& 60.1&57.6&34.3 & 41.4&\underline{39.8}&38.5 & 6.7\\ % \hline
 MapTRv2\cite{liao2023maptrv2}& R50& 24  &50& 59.8 &62.4& 62.4& 61.5& 34.1\dag& 41.6\dag&  36.8\dag&37.5\dag &11.6\\
 % DTCLMapper\cite{li2024dtclmapper}& R50& 24 &50& 59.7& 62.6& 63.4& 61.9& -& -& -& -\\
 MGMap\cite{liu2024mgmap}& R50& 24  &50& \underline{61.8} & \textbf{65.0}& \textbf{67.5}& \underline{64.8}&  -&  -&   -&  -&-\\
  \textbf{PerCMap}*& R50& 24 &50& \textbf{64.8}& \underline{63.6}&\underline{66.8}& \textbf{65.1}& \textbf{39.4}&\underline{43.9}& \textbf{41.8}&\textbf{41.7}&10.5\\
% PerCMap& R50& 24 &80& 64.9&65.9 &68.8&66.5&41.3&46.6& 43.7&43.9\\ 
\hline
 MapQR\cite{liu2024mapqr}& R50& 24 &100& 63.4& \textbf{68.0}& 67.7& 66.4& \underline{38.6}& \underline{49.9}& \underline{41.5}&\underline{43.3}&10.9\\ 
  MGMapNet\cite{yang2024mgmapnet}& R50& 24& 100& \underline{64.7}& 66.1& \textbf{69.4}& \underline{66.8}& -& -& -&-&-\\
         
 % MapTR~\shortcite{liao2022maptr}& R50  & 110   & 56.2& 59.8& 60.1 & 58.7 &  -& -& -& -\\
 % BeMapNet~\shortcite{qiao2023bemapnet}& Swin-T& 110& 61.3& 64.4& 61.6& 62.5& 42.2& 49.1& 39.9&43.7\\
 % MapTRv2~\shortcite{liao2023maptrv2}& R50& 110 & 68.1 & 68.3 & 69.7 & 68.7& -& -&- &- \\
 % HybriMap& R50 & 110   & \textbf{68.8}& \textbf{68.5}&\textbf{71.1} & \textbf{69.4}&  \textbf{46.1}& \textbf{50.6} & \textbf{47.6}& \textbf{48.1} \\
 %        \hline
 \textbf{PerCMap-QR}& R50& 24& 100& \textbf{65.6}& \underline{67.5}& \underline{68.2}& \textbf{67.1}& \textbf{39.1}& \textbf{50.0}& \textbf{43.4}&\textbf{45.0}&8.7\\ \hline
    \end{tabular} % \vspace{-18pt}

    \label{tab:bs_nusc}
\end{table*}

\begin{table}[t]
\setlength{\tabcolsep}{3pt}
    \caption{Comparison of 2D and 3D vectorized map construction on \textit{Argoverse 2} val set. 
    Top and middle sections show 2D HD map construction; bottom section shows 3D vector predictions.}
    \centering
    \label{tab:bs_argo}
    \begin{tabular}{c|lc|cccc}
    \hline
    & Method & \(N_m\) & \(\mbox{AP}_{ped}\) & \(\mbox{AP}_{div}\) & \(\mbox{AP}_{bdr}\) & mAP \\
    \hline

\multirow{5}{*}{
  \rotatebox{90}{
    \parbox{1.3cm}{\centering Coarse\\thr.\\(2D)}
  }
}
    & VectormapNet\cite{liu2023vectormapnet} & -- & 38.3 & 36.1 & 39.2 & 37.9 \\
    & MGMap\cite{liu2024mgmap} & 50 & 52.8 & 67.5 & 68.1 & 62.8 \\
    & MapTRv2\cite{liao2023maptrv2} & 50 & 62.9 & 72.1 & 67.1 & 67.4 \\
    & HIMap\cite{map:zhou2024himap} & -- & \textbf{69.0} & 69.5 & 70.3 & 69.6 \\
    & \textbf{PerCMap} & 50 & 67.4 & \textbf{73.0} & \textbf{71.1} & \textbf{70.5} \\
    \hline
\multirow{3}{*}{
  \rotatebox{90}{
    \parbox{0.8cm}{\centering Tight\\thr. \\(2D)}
  }
}
    & PivotNet\cite{ding2023pivotnet} & 60 & 31.3 & 47.5 & 43.4 & 40.7 \\
    & HIMap\cite{map:zhou2024himap} & -- & 39.9 & 53.4 & 44.3 & 45.8 \\
    & \textbf{PerCMap} & 50 & \textbf{38.7} & \textbf{55.7} & \textbf{46.0} & \textbf{46.8} \\
    \hline

\multirow{3}{*}{
  \rotatebox{90}{
    \parbox{0.8cm}{\centering 3D map \\pred.}
  }
}
    & VectormapNet\cite{liu2023vectormapnet} & -- & 36.5 & 35.0 & 36.2 & 35.8 \\
    & MapTRv2\cite{liao2023maptrv2} & 50 & 60.7 & 68.9 & 64.5 & 64.7 \\
    & \textbf{PerCMap} & 50 & \textbf{64.6} & \textbf{71.2} & \textbf{69.6} & \textbf{68.4} \\
    \hline
    \end{tabular}
    % \vspace{-16pt}
\end{table}

\section{Experiments}

\subsection{Experimental Settings}

\subsubsection{Datasets}

We evaluate our method on two popular benchmarks: \textit{nuScenes}\cite{caesar2020nuscenes}  and \textit{Argoverse 2}~\cite{Argoverse2}. Each dataset comprises 1000 scenes in total, which are divided into train/validation/test sets with a split of 700/150/150 scenes. In \textit{nuScenes}, each scene contains a video approximately 20s long, where key frames are annotated at 2Hz. These key frames are all captured by 6 surround view onboard cameras with 360 degrees field-of-view, containing 6 images for each. \textit{Argoverse 2} dataset annotates key frames at 10Hz, in which the key frame contains 7 ring cameras.
\subsubsection{Metric}
Following the previous works\cite{liu2023vectormapnet,liao2022maptr}, we use Average Precision (AP) to evaluate the performance of the methods. A predicted instance is defined as true positive only if its Chamfer Distance from the groundtruth is less than a certain threshold. By default, thresholds are set to \(\{0.5m, 1.0m, 1.5m\}\), which we refer to as \textbf{coarse thresholds}. To enable more fine-grained analysis, we additionally report results using \textbf{tight thresholds} \(\{0.2m, 0.5m, 1.0m\}\), following suggestions in \cite{qiao2023bemapnet,ding2023pivotnet}. We denote the mean AP over coarse and tight thresholds as \textbf{\(\mbox{mAP}\)}, and \textbf{\(\mbox{mAP}_{T}\)}, respectively. 
\subsubsection{Implementation Details}
The map instances involve static elements including pedestrian crossing (\textit{ped}), lane divider (\textit{div}), and road boundary (\textit{bdr}). The perceptual range spans \(15m\) laterally and \(30m\) longitudinally around the ego vehicle, with a BEV feature resolution of \(100\times200\). Each predicted vector contains 20 points. In the Cross-view Instance Activation (CIA) module, the number of activated instances per single view is set to \(N_m^P=25\). The corresponding loss weights are set to \(\gamma_H=0.1\) and \(\gamma_{RIS}=15\). 

We conduct experiments on both the MapTRv2~\cite{liao2023maptrv2} and MapQR~\cite{liu2024mapqr} baselines, which are denoted as PerCMap and PerCMap-QR, respectively, in the following results. For both settings, we adopt ResNet-50 (R50)\cite{he2016deep} as the image backbone, while keeping the same PV-to-BEV transformation module and map decoder as in the corresponding baselines to ensure a fair comparison. The number of predicted map instances is set to \(N_m=50\) for PerCMap and \(N_m=100\) for PerCMap-QR. The AdamW optimizer\cite{loshchilov2018decoupled} is employed for all experiments. An initial learning rate of \(6\times10^{-4}\) is used in all experiments, including in ablation studies, except for the PerCMap-QR variant, where a learning rate of \(5\times10^{-4}\) is adopted to ensure training stability.

\subsection{Main Results}
\subsubsection{Quantitative Results on \textit{nuScenes} }
We evaluate the performance of PerCMap on the \textit{nuScenes} validation set and compare it with existing methods, as shown in Tab. \ref{tab:bs_nusc}. Using 50 predicted vectors, PerCMap achieves mAP scores of 65.1 and 41.7 under the coarse and tight thresholds, respectively, surpassing the baseline MapTRv2\cite{liao2023maptrv2} by 3.6 and 4.2 mAP. The PerCMap-QR variant further improves performance, reaching 67.1 and 45.0 mAP on the two threshold sets. Such improvements are achieved with only 1.1 and 2.2 FPS reduction. Taking the tight threshold as an example, this corresponds to merely 0.26 and 1.3 FPS per unit mAP improvement for PerCMap and PerCMap-QR, respectively. Considering the influence of the increased instance amount on network scale in the latter case, the overall runtime cost remains acceptable.

Notably, our method consistently performs better under the tight threshold setting, indicating its effectiveness in accurately decoding the fine-grained geometry of map vectors. PerCMap exhibits improved stability in detecting pedestrian crossings and road boundaries, especially under tight thresholds. These categories typically present more complex shapes and stronger perspective-view priors than lane dividers. This demonstrates the model’s enhanced capability to localize points precisely by effectively leveraging perspective-view cues.

\subsubsection{Quantitative Results on \textit{Argoverse 2}. }
To further demonstrate the effectiveness of our approach, we present results on the larger \textit{Argoverse 2} dataset (Tab. \ref{tab:bs_argo}). All experiments were conducted over 6 epoch training. PerCMap achieves 70.5 and 46.8 mAP under two thresholds, despite using an equal or smaller number of predicted map instances. In addition, PerCMap shows advantages in constructing vectorized 3D maps by lifting map vectors into 3D space.  For clarity, we refer to HD map construction in the BEV space as the 2D case. The results in both 2D and 3D settings significantly outperform those of previous methods.

\subsubsection{Qualitative Analysis}
\begin{figure}[t]
        \centering
    \includegraphics[width=0.98\linewidth]{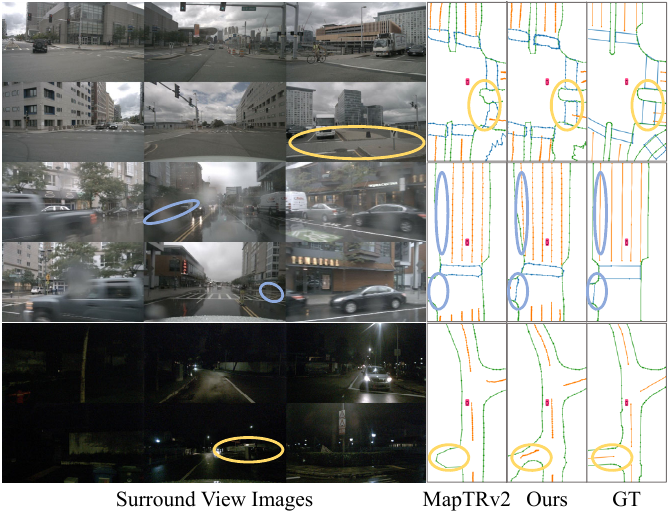} %  \vspace{-8pt}
    \caption{\textbf{Qualitative results on \textit{nuScenes} validation dataset.} We provide the complete surround view inputs, predictions of MapTRv2\protect\cite{liao2023maptrv2}, and PerCMap, and the Ground Truth map. The regions that are highlighted by colored ellipses enclose instances that are difficult to detect. } % \vspace{-10pt}
    \label{fig:nusc}
\end{figure}
\begin{figure}[]
        \centering
    \includegraphics[width=0.9\linewidth]{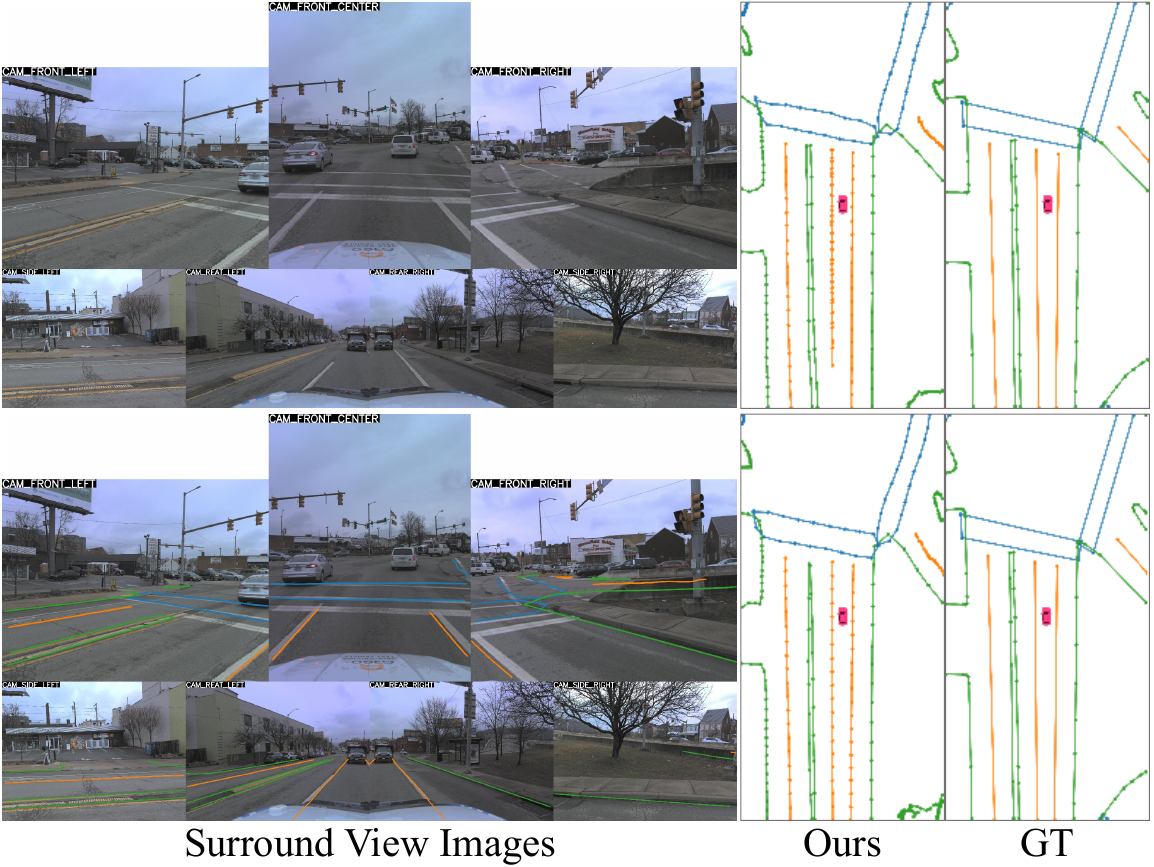} % \vspace{-8pt}
    \caption{\textbf{Qualitative results on \textit{Argoverse 2} validation dataset.} The \textbf{upper} line displays input and output samples 2D from HD map construction in BEV, and the \textbf{bottom} line showcases the results of 3D map construction, with map vectors projected onto PV and BEV.  } % \vspace{-22pt}
    % PerCMap demonstrates strong predictive performance in both tasks.
    \label{fig:argo}
\end{figure}

We present the visualized map construction results under various weather or illumination conditions in both datasets. Fig. \ref{fig:nusc} includes the comparative examples from \textit{nuScenes}, while Fig. \ref{fig:argo} includes cases from \textit{Argoverse 2}. In Fig. \ref{fig:nusc}, we specifically emphasize the areas of map elements that are difficult to detect. Compared to MapTRv2, PerCMap successfully detects distant crossings and blurred lanes present in the original image in the case of Row 2. It also provides more accurate shape predictions for road boundaries or crossings in the cases of Row 1 and Row 3. Overall, the approach of emphasizing instance features and optimizing point decoding based on the input images are effective. Additionally, the results in the left section of Fig. \ref{fig:argo} show that PerCMap achieves strong performance in detecting map elements in \textit{Argoverse 2}. 

Nonetheless, our method’s performance is constrained when PV features become difficult to extract due to poor visibility. For instance, in Row 3 of Fig. \ref{fig:nusc}, the sidewalk is mispredicted with a noticeable directional offset, likely caused by missing or ambiguous cues in the original image.

\subsection{Ablation Studies}
In this section, we present ablation studies to evaluate the effectiveness of key components, including CIA, DPE, \(\mathcal{L}_{RIS}\), and specific designs within CIA and DPE. We also provide comparisons between PerCMap and the baseline under different PV-to-BEV methods, various weather conditions, and more challenging dataset splits. Unless otherwise specified, all ablation settings follow the PerCMap configuration in Tab.\ref{tab:bs_nusc}
\subsubsection{Ablation on Proposed Modules}
To deeply investigate their effectiveness in map prediction, we conduct ablation experiments on CIA,  DPE and \(\mathcal{L}_{RIS}\), and list the experimental results in Tab. \ref{tab:inspts_new}. Each module contributes positively to overall performance. Compared to the baseline, introducing CIA (Row 2) or DPE (Row 3) individually yields notable improvements of 1.8 and 2.5 \(\mbox{mAP}\). When used together, yield a 3.0 gain in \(\mbox{mAP}\). The proposed supervision loss \(\mathcal{L}_{RIS}\) provides gain of 0.5 \(\mbox{mAP}\) and 1.0 \(\mbox{mAP}_{T}\), effectively constraining the geometric accuracy of map vectors. 

\begin{table}
\setlength{\tabcolsep}{3pt}
    \caption{Ablation results on CIA and DPE modules.}
    \centering
    \begin{tabular}{c|ccc|cccc|c}
        \hline
           No.&CIA&DPE &\(\mathcal{L}_{RIS}\)& \(\mbox{AP}_{ped}\) & \(\mbox{AP}_{div}\) & \(\mbox{AP}_{bdr}\) &  \(\mbox{mAP}\) &\(\mbox{mAP}_{T}\)\\         \hline
 1& & & & 59.8& 62.4& 62.4&61.5 &37.5\\

           2&\Checkmark& 
&
& 61.8& 62.7& 65.5& 63.3&39.1\\
   3&&  \Checkmark 
&& 62.8& 64.2& 65.1&64.0&39.8\\
   4&\Checkmark&  \Checkmark && 62.9& 64.7& 66.0&64.5&40.6\\ \hline
 5& & & \Checkmark
& 60.5& 62.2& 63.4& 62.0&38.5\\
        
   6&
\Checkmark& \Checkmark &\Checkmark
& 64.8& 63.6& 66.8&65.1 &41.7\\\hline
    \end{tabular} % \vspace{-12pt}

    \label{tab:inspts_new}
\end{table}

\subsubsection{Analysis on CIA Design}
In CIA, PV features are converted into \(N_m^P\) instance features, each serving as an activated query. Thus, \(N_m^P\) controls the conversion scale from surround-view features to instance-level representations. As shown in Tab. \ref{tab:cia}, with the total predicted vectors fixed at \(N_m=50\), too small \(N_m^P\) causes insufficient activation, while too large introduces excessive parameters that hinder learning efficiency. The best performance is achieved at  \(N_m^P=25\).

\begin{table}
% \color{red}
    \caption{Results on different value about \(N_m^P\).}
    \centering
    \begin{tabular}{c|cc|c|cc}
        \hline
         \(N_m^P\)&  \(\mbox{mAP}\) &\(\mbox{mAP}_{T}\) &  \(N_m^P\)& \(\mbox{mAP}\) &\(\mbox{mAP}_{T}\) \\
        \hline
         5& 62.5& 38.3&  15& 62.6&38.4\\
        
  25&63.3& 39.1&  40& 62.9&38.4\\\hline
    \end{tabular} % \vspace{-16pt}

    \label{tab:cia}
\end{table}

\subsubsection{Analysis on DPE Design}

\begin{figure*}[t]
    \centering
    \includegraphics[width=0.9\linewidth]{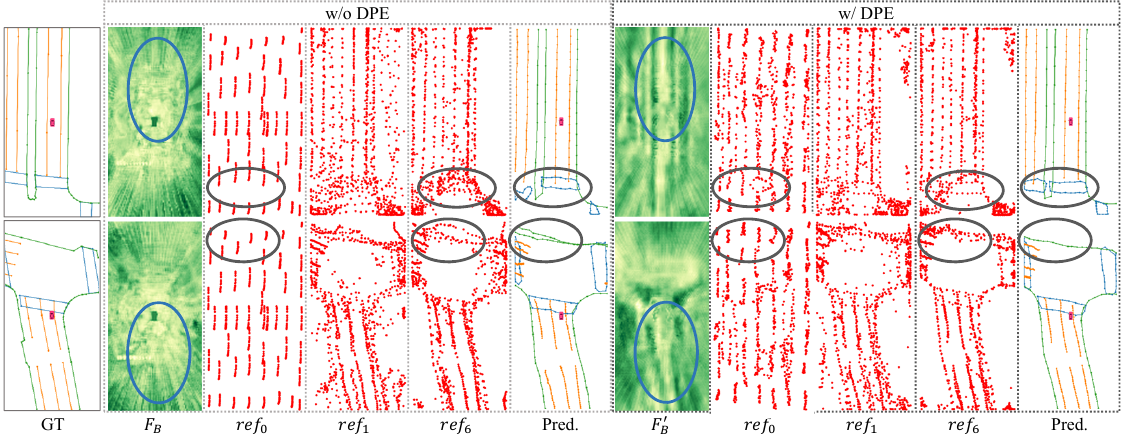} \vspace{-10pt}
    \caption{\textbf{Illustration of reference points and BEV Features w/o and w/ DPE.} ``\(ref_{x}\)" represents the reference points after the \(x\)-th layer, while  ``\(ref_{0}\)" refers to the initialized points without being processed by any decoder layer. With the inclusion of DPE, the reference points tend to align more closely with map elements, and the BEV features are more capable of accurately represent map information within the scene. } %  \vspace{-16pt}
    \label{fig:feats}
\end{figure*}
DPE module consists of three components: positional embedding integration, heatmap loss, and BEV feature enhancement. Since the effectiveness of BEV enhancement relies on supervision from the heatmap branch, we incorporate these components sequentially and evaluate their respective contributions. The results in Tab. \ref{tab:dpe_new} show that each branch improves accuracy. In particular, the heatmap loss yields notable improvements under tight threshold metrics. We attribute this to the effective supervision of keypoints, enabling more precise reconstruction of map shapes.

To illustrate its functionality, we present visualization results of DPE in Fig. \ref{fig:feats}. Since positional embedding vectors are difficult to plot, we instead use reference points initialized from Integrated Positional Embeddings. The initialized reference points drift toward map regions compared with random initialization. With DPE, the reference points become input-dependent and converge around map elements in later iterations as shown in the gray circles. Meanwhile, DPE also improves the BEV features' ability to represent the scene's map information as shown in the blue circles . 

\begin{table}
\setlength{\tabcolsep}{4pt}

    \caption{Results on the components in DPE. }
    \centering
    \begin{tabular}{l|cccc|c}
        \hline
         Method& \(\mbox{AP}_{ped}\) & \(\mbox{AP}_{div}\) & \(\mbox{AP}_{bdr}\) &  \(\mbox{mAP}\) &\(\mbox{mAP}_{T}\)\\ \hline
  Baseline& 59.8& 62.4& 62.4&61.5 &37.5\\
    + Int. Pos. Embeddings& 59.1& 63.5& 64.8& 62.5&38.1\\
   + Heatmap Loss \(\mathcal{L}_{H}\)& 61.2& 63.1& 64.9& 63.1&39.2\\ 
           + BEV Enhancement& 62.8& 64.2&65.1& 64.0&39.8\\
        \hline
    \end{tabular} %  \vspace{-12pt}

    \label{tab:dpe_new}
\end{table}

\subsubsection{Robustness on PV-to-BEV Modules}
PerCMap enhances map construction by mitigating information loss in PV-to-BEV transformation. As shown in Tab. \ref{tab:viewtr}, we compare it with baseline under different view transformation methods, including LSS (with and without depth supervision) and GKT \cite{chen2022gkt}. The results show that PerCMap is compatible with various PV-to-BEV modules and consistently improves performance.

\begin{table}
% \arrayrulecolor{red}
% \color{red}    
    \caption{Comparison results on different PV-to-BEV transformation.}
    \centering
    \begin{tabular}{l|cc|cc}        \hline
 \multirow{ 2}{*}{PV-to-BEV}&  \multicolumn{2}{c|}{MapTRv2}& \multicolumn{2}{c}{PerCMap}  \\

        & \(\mbox{mAP}\)&  \(\mbox{mAP}_{T}\) &\(\mbox{mAP}\)&  \(\mbox{mAP}_{T}\)\\
        \hline
        GKT&55.2&  31.8& 59.0(+3.8\(\uparrow\))&36.3(+4.5\(\uparrow\))\\ 
 LSS&58.0&  34.6& 62.5(+4.4\(\uparrow\))&39.3(+4.7\(\uparrow\))\\  
  LSS-depth&61.5& 37.5 & 65.1(+3.6\(\uparrow\))& 41.7(+4.2\(\uparrow\))\\ \hline
    \end{tabular} %  \vspace{-14pt}

    \label{tab:viewtr}
\end{table}

\subsubsection{Robustness on Weather and Lighting}
To assess robustness under different weather and lighting conditions, we compare our method with MapTRv2 in sunny, rainy, daytime, and nighttime scenes (Tab.~\ref{tab:weather}), following the official \textit{nuScenes} split. Our network achieves notable gains in sunny, rainy, and daytime scenarios, and still improves by about 1.0 mAP at night despite reduced visibility. Overall, it consistently delivers improvements across diverse conditions.
\begin{table}
% \arrayrulecolor{red}
% \color{red}    
% \vspace{-6pt}
    \caption{Comparison results on various weather and lighting.}
    \centering
    \begin{tabular}{l|cc|cc}        \hline
 \multirow{ 2}{*}{Scene}&  \multicolumn{2}{c|}{MapTRv2}& \multicolumn{2}{c}{PerCMap}  \\

        & \(\mbox{mAP}\)&  \(\mbox{mAP}_{T}\) &\(\mbox{mAP}\)&  \(\mbox{mAP}_{T}\)\\
        \hline
        Sunny&63.6&  39.1& 67.1(+3.5\(\uparrow\))&43.4(+4.3\(\uparrow\))\\ 
 Rainy&52.1&  30.1& 55.1(+3.0\(\uparrow\))&33.6(+3.5\(\uparrow\))\\
 Day& 62.4& 38.1& 66.0(+3.6\(\uparrow\))&42.4(+4.3\(\uparrow\))\\  
  Night&39.5& 21.1& 40.4(+0.9\(\uparrow\))& 23.1(+1.0\(\uparrow\))\\ \hline
    \end{tabular} 

    \label{tab:weather}
\end{table}

\subsubsection{Results on Geospatial Disjoint Split}
We evaluate on the geospatial disjoint split \cite{yuan2024streammapnet} of \textit{nuScenes}, which reduces overlap between training and validation sets and thus increases inference difficulty. Compared with the baseline, our method achieves a 2.2 \(\mbox{mAP}\) gain under the tight threshold criterion.
\begin{table}
    \caption{Results on geospatial disjoint split of \textit{nuScenes}.}
    \centering
    \begin{tabular}{l|cc|l|cc} 
    \hline
         Method&  \(\mbox{mAP}\) &\(\mbox{mAP}_{T}\) &  Method&\(\mbox{mAP}\) &\(\mbox{mAP}_{T}\) \\ \hline
         MapTRv2& 28.0& 13.2&  PerCMap&29.5(+1.5\(\uparrow\))&15.4(+2.2\(\uparrow\))\\ \hline
     \end{tabular} % \vspace{-12pt}
 \label{tab:newsplit}
\end{table}

\subsubsection{Additional analysis on Mask2Map}

A recent study, Mask2Map~\cite{choi2024mask2map}, formulates HD map construction as a two-phase framework. In the first phase, it trains a BEV instance segmentor, and in the second phase it exploits the extracted geometrical priors for map decoding. Our proposed PV reuse mechanism and RIS loss supervision are also effective within this two-phase setting, where a well-trained BEV representation is available. As a validation, we conduct experiments using Mask2Map as the baseline, with results summarized in Tab.~\ref{tab:mask2map}. In particular, the experiments with +PerCMap correspond to further adding the CIA and DPE modules on top of +RIS loss. Taking the coarse thresholds as an example, the inclusion of RIS loss and PerCMap leads to improvements of 1.0 and 1.5 mAP, respectively. Since the baseline already employs query and positional embedding adaptation as well as supervised BEV representations, the feature encoding is relatively strong. Nevertheless, our modules still provide consistent performance gains.

\begin{table}
% \color{red} 
% \vspace{-4pt}
    \caption{Results on Mask2Map baseline.}
    \centering
    \begin{tabular}{l|cc} 
    \hline
         Experiment&  \(\mbox{mAP}\)&\(\mbox{mAP}_{T}\)\\ \hline
         Mak2Map& 71.5 &49.9\\
         Mak2Map+RIS Loss& 
    72.5(+1.0\(\uparrow\))&51.8(+1.9\(\uparrow\))\\ 
 Mak2Map+PerCMap&  73.0(+1.5\(\uparrow\))&52.1(+2.2\(\uparrow\))\\ \hline\end{tabular} % \vspace{-16pt}
    \label{tab:mask2map}
\end{table}

\section{Conclusion}
In this paper, we propose PerCMap to improve the performance of HD map construction from instance and point level. We exploit the feature of the inputs, i.e. perspective-view image features. Specifically, we build the Cross-view Instance Activation to emphasize the instance feature; meanwhile, we design the Dual-view Point Embedding to strengthen the map vector decoding at point level. Extensive experiments on the dataset have demonstrated the superiority of our method, achieving appreciable performance in HD map construction. We anticipate that the aforementioned research can provide more reliable map support for autonomous driving and contribute to future investigations in HD map studies.

% or


% \appendices

% % you can choose not to have a title for an appendix
% % if you want by leaving the argument blank
% \section{}

% % use section* for acknowledgment
% \section*{Acknowledgment}


\ifCLASSOPTIONcaptionsoff
  \newpage
\fi



%\bibliographystyle{IEEEtran}


% \end{thebibliography}
\bibliographystyle{IEEEtran}
\bibliography{myieee}

\begin{thebibliography}{10}
\providecommand{\url}[1]{#1}
\csname url@samestyle\endcsname
\providecommand{\newblock}{\relax}
\providecommand{\bibinfo}[2]{#2}
\providecommand{\BIBentrySTDinterwordspacing}{\spaceskip=0pt\relax}
\providecommand{\BIBentryALTinterwordstretchfactor}{4}
\providecommand{\BIBentryALTinterwordspacing}{\spaceskip=\fontdimen2\font plus
\BIBentryALTinterwordstretchfactor\fontdimen3\font minus \fontdimen4\font\relax}
\providecommand{\BIBforeignlanguage}[2]{{%
\expandafter\ifx\csname l@#1\endcsname\relax
\typeout{** WARNING: IEEEtran.bst: No hyphenation pattern has been}%
\typeout{** loaded for the language `#1'. Using the pattern for}%
\typeout{** the default language instead.}%
\else
\language=\csname l@#1\endcsname
\fi
#2}}
\providecommand{\BIBdecl}{\relax}
\BIBdecl

\bibitem{liao2023maptrv2}
B.~Liao, S.~Chen, Y.~Zhang, B.~Jiang, Q.~Zhang, W.~Liu, C.~Huang, and X.~Wang, ``Maptrv2: An end-to-end framework for online vectorized hd map construction,'' \emph{arXiv preprint arXiv:2308.05736}, 2023.

\bibitem{lu_monocular_2019}
\BIBentryALTinterwordspacing
C.~Lu, M.~J.~G. van~de Molengraft, and G.~Dubbelman, ``\BIBforeignlanguage{en}{Monocular {Semantic} {Occupancy} {Grid} {Mapping} with {Convolutional} {Variational} {Encoder}-{Decoder} {Networks}},'' \emph{\BIBforeignlanguage{en}{IEEE Robotics and Automation Letters}}, vol.~4, no.~2, pp. 445--452, Apr. 2019, arXiv:1804.02176 [cs]. [Online]. Available: \url{http://arxiv.org/abs/1804.02176}
\BIBentrySTDinterwordspacing

\bibitem{roddick_predicting_2020}
\BIBentryALTinterwordspacing
T.~Roddick and R.~Cipolla, ``\BIBforeignlanguage{en}{Predicting {Semantic} {Map} {Representations} {From} {Images} {Using} {Pyramid} {Occupancy} {Networks}},'' in \emph{\BIBforeignlanguage{en}{2020 {IEEE}/{CVF} {Conference} on {Computer} {Vision} and {Pattern} {Recognition} ({CVPR})}}.\hskip 1em plus 0.5em minus 0.4em\relax Seattle, WA, USA: IEEE, Jun. 2020, pp. 11\,135--11\,144. [Online]. Available: \url{https://ieeexplore.ieee.org/document/9156806/}
\BIBentrySTDinterwordspacing

\bibitem{yang2021pyva}
W.~Yang, Q.~Li, W.~Liu, Y.~Yu, Y.~Ma, S.~He, and J.~Pan, ``Projecting your view attentively: Monocular road scene layout estimation via cross-view transformation,'' in \emph{Proceedings of the IEEE/CVF conference on computer vision and pattern recognition}, 2021, pp. 15\,536--15\,545.

\bibitem{zhou2022cross}
B.~Zhou and P.~Kr{\"a}henb{\"u}hl, ``Cross-view transformers for real-time map-view semantic segmentation,'' in \emph{Proceedings of the IEEE/CVF conference on computer vision and pattern recognition}, 2022, pp. 13\,760--13\,769.

\bibitem{li2022hdmapnet}
Q.~Li, Y.~Wang, Y.~Wang, and H.~Zhao, ``Hdmapnet: An online hd map construction and evaluation framework,'' in \emph{2022 International Conference on Robotics and Automation (ICRA)}.\hskip 1em plus 0.5em minus 0.4em\relax IEEE, 2022, pp. 4628--4634.

\bibitem{liu2023vectormapnet}
Y.~Liu, T.~Yuan, Y.~Wang, Y.~Wang, and H.~Zhao, ``Vectormapnet: End-to-end vectorized hd map learning,'' in \emph{International Conference on Machine Learning}.\hskip 1em plus 0.5em minus 0.4em\relax PMLR, 2023, pp. 22\,352--22\,369.

\bibitem{liao2022maptr}
B.~Liao, S.~Chen, X.~Wang, T.~Cheng, Q.~Zhang, W.~Liu, and C.~Huang, ``Maptr: Structured modeling and learning for online vectorized hd map construction,'' in \emph{The Eleventh International Conference on Learning Representations}, 2022.

\bibitem{ding2023pivotnet}
W.~Ding, L.~Qiao, X.~Qiu, and C.~Zhang, ``Pivotnet: Vectorized pivot learning for end-to-end hd map construction,'' in \emph{Proceedings of the IEEE/CVF International Conference on Computer Vision (ICCV)}, October 2023, pp. 3672--3682.

\bibitem{map:zhou2024himap}
Y.~Zhou, H.~Zhang, J.~Yu, Y.~Yang, S.~Jung, S.-I. Park, and B.~Yoo, ``Himap: Hybrid representation learning for end-to-end vectorized hd map construction,'' in \emph{Proceedings of the IEEE/CVF Conference on Computer Vision and Pattern Recognition}, 2024, pp. 15\,396--15\,406.

\bibitem{choi2024mask2map}
S.~Choi, J.~Kim, H.~Shin, and J.~W. Choi, ``Mask2map: Vectorized hd map construction using bird’s eye view segmentation masks,'' in \emph{European Conference on Computer Vision}.\hskip 1em plus 0.5em minus 0.4em\relax Springer, 2024, pp. 19--36.

\bibitem{yang2024mgmapnet}
J.~Yang, M.~Jiang, S.~Yang, X.~Tan, Y.~Li, E.~Ding, H.~Wang, and J.~Wang, ``Mgmapnet: Multi-granularity representation learning for end-to-end vectorized hd map construction,'' \emph{arXiv preprint arXiv:2410.07733}, 2024.

\bibitem{yang2023bevformerv2}
C.~Yang, Y.~Chen, H.~Tian, C.~Tao, X.~Zhu, Z.~Zhang, G.~Huang, H.~Li, Y.~Qiao, L.~Lu \emph{et~al.}, ``Bevformer v2: Adapting modern image backbones to bird's-eye-view recognition via perspective supervision,'' in \emph{Proceedings of the IEEE/CVF Conference on Computer Vision and Pattern Recognition}, 2023, pp. 17\,830--17\,839.

\bibitem{song2024sdtr}
Q.~Song, Q.~Hu, C.~Zhang, Y.~Chen, and R.~Huang, ``Divide and conquer: Improving multi-camera 3d perception with 2d semantic-depth priors and input-dependent queries,'' \emph{IEEE Transactions on Image Processing}, 2024.

\bibitem{liu2024mgmap}
X.~Liu, S.~Wang, W.~Li, R.~Yang, J.~Chen, and J.~Zhu, ``Mgmap: Mask-guided learning for online vectorized hd map construction,'' in \emph{Proceedings of the IEEE/CVF Conference on Computer Vision and Pattern Recognition}, 2024, pp. 14\,812--14\,821.

\bibitem{liu2024mapqr}
Z.~Liu, X.~Zhang, G.~Liu, J.~Zhao, and N.~Xu, ``Leveraging enhanced queries of point sets for vectorized map construction,'' in \emph{European Conference on Computer Vision}, 2024.

\bibitem{qiu2024ticmapnet}
W.~Qiu, S.~Pang, H.~Zhang, J.~Fang, and J.~Xue, ``Ticmapnet: A tightly coupled temporal fusion pipeline for vectorized hd map learning,'' \emph{IEEE Robotics and Automation Letters}, 2024.

\bibitem{wang2022detr3d}
Y.~Wang, V.~C. Guizilini, T.~Zhang, Y.~Wang, H.~Zhao, and J.~Solomon, ``Detr3d: 3d object detection from multi-view images via 3d-to-2d queries,'' in \emph{Conference on Robot Learning}.\hskip 1em plus 0.5em minus 0.4em\relax PMLR, 2022, pp. 180--191.

\bibitem{philion2020lift}
J.~Philion and S.~Fidler, ``Lift, splat, shoot: Encoding images from arbitrary camera rigs by implicitly unprojecting to 3d,'' in \emph{Proceedings of the European Conference on Computer Vision}, 2020.

\bibitem{li2022bevformer}
Z.~Li, W.~Wang, H.~Li, E.~Xie, C.~Sima, T.~Lu, Y.~Qiao, and J.~Dai, ``Bevformer: Learning bird’s-eye-view representation from multi-camera images via spatiotemporal transformers,'' in \emph{European conference on computer vision}.\hskip 1em plus 0.5em minus 0.4em\relax Springer, 2022, pp. 1--18.

\bibitem{liu2022petr}
Y.~Liu, T.~Wang, X.~Zhang, and J.~Sun, ``Petr: Position embedding transformation for multi-view 3d object detection,'' in \emph{European Conference on Computer Vision}.\hskip 1em plus 0.5em minus 0.4em\relax Springer, 2022, pp. 531--548.

\bibitem{zou2023hft}
J.~Zou, Z.~Zhu, J.~Huang, T.~Yang, G.~Huang, and X.~Wang, ``Hft: Lifting perspective representations via hybrid feature transformation for bev perception,'' in \emph{2023 IEEE International Conference on Robotics and Automation (ICRA)}.\hskip 1em plus 0.5em minus 0.4em\relax IEEE, 2023, pp. 7046--7053.

\bibitem{li2023bimapper}
S.~Li, K.~Yang, H.~Shi, J.~Zhang, J.~Lin, Z.~Teng, and Z.~Li, ``Bi-mapper: Holistic bev semantic mapping for autonomous driving,'' \emph{IEEE Robotics and Automation Letters}, 2023.

\bibitem{huang2025duospacenet}
Z.~Huang, Y.~Zhao, H.~Xiao, C.~Wu, and L.~Ge, ``Duospacenet: Leveraging both bird's-eye-view and perspective view representations for 3d object detection,'' in \emph{Proceedings of the Computer Vision and Pattern Recognition Conference}, 2025, pp. 2560--2570.

\bibitem{park2023solofusion}
J.~Park, C.~Xu, S.~Yang, K.~Keutzer, K.~M. Kitani, M.~Tomizuka, and W.~Zhan, ``Time will tell: New outlooks and a baseline for temporal multi-view 3d object detection,'' in \emph{ICLR}, 2023.

\bibitem{monninger2024tempbev}
T.~Monninger, V.~Dokkadi, M.~Z. Anwar, and S.~Staab, ``Tempbev: Improving learned bev encoders with combined image and bev space temporal aggregation,'' in \emph{2024 IEEE/RSJ International Conference on Intelligent Robots and Systems (IROS)}.\hskip 1em plus 0.5em minus 0.4em\relax IEEE, 2024, pp. 9668--9675.

\bibitem{huang2024anchor3dlane++}
S.~Huang, Z.~Shen, Z.~Huang, Y.~Liao, J.~Han, N.~Wang, and S.~Liu, ``Anchor3dlane++: 3d lane detection via sample-adaptive sparse 3d anchor regression,'' \emph{IEEE Transactions on Pattern Analysis and Machine Intelligence}, 2024.

\bibitem{qiao2023bemapnet}
L.~Qiao, W.~Ding, X.~Qiu, and C.~Zhang, ``End-to-end vectorized hd-map construction with piecewise bezier curve,'' in \emph{Proceedings of the IEEE/CVF Conference on Computer Vision and Pattern Recognition}, 2023, pp. 13\,218--13\,228.

\bibitem{cheng2022sparse}
T.~Cheng, X.~Wang, S.~Chen, W.~Zhang, Q.~Zhang, C.~Huang, Z.~Zhang, and W.~Liu, ``Sparse instance activation for real-time instance segmentation,'' in \emph{Proceedings of the IEEE/CVF Conference on Computer Vision and Pattern Recognition}, 2022, pp. 4433--4442.

\bibitem{luo2023latr}
Y.~Luo, C.~Zheng, X.~Yan, T.~Kun, C.~Zheng, S.~Cui, and Z.~Li, ``Latr: 3d lane detection from monocular images with transformer,'' in \emph{Proceedings of the IEEE/CVF International Conference on Computer Vision}, 2023, pp. 7941--7952.

\bibitem{det:zhu2020deformable}
X.~Zhu, W.~Su, L.~Lu, B.~Li, X.~Wang, and J.~Dai, ``Deformable detr: Deformable transformers for end-to-end object detection,'' in \emph{International Conference on Learning Representations}, 2020.

\bibitem{det:law2018cornernet}
H.~Law and J.~Deng, ``Cornernet: Detecting objects as paired keypoints,'' in \emph{Proceedings of the European conference on computer vision (ECCV)}, 2018, pp. 734--750.

\bibitem{det:duan2019centernet}
K.~Duan, S.~Bai, L.~Xie, H.~Qi, Q.~Huang, and Q.~Tian, ``Centernet: Keypoint triplets for object detection,'' in \emph{Proceedings of the IEEE/CVF international conference on computer vision}, 2019, pp. 6569--6578.

\bibitem{det:lin2017focal}
T.-Y. Lin, P.~Goyal, R.~Girshick, K.~He, and P.~Doll{\'a}r, ``Focal loss for dense object detection,'' in \emph{Proceedings of the IEEE international conference on computer vision}, 2017, pp. 2980--2988.

\bibitem{efficientnet2019}
\BIBentryALTinterwordspacing
M.~Tan and Q.~Le, ``{EfficientNet}: {Rethinking} {Model} {Scaling} for {Convolutional} {Neural} {Networks},'' in \emph{Proceedings of the 36th {International} {Conference} on {Machine} {Learning}}, ser. Proceedings of {Machine} {Learning} {Research}, K.~Chaudhuri and R.~Salakhutdinov, Eds., vol.~97.\hskip 1em plus 0.5em minus 0.4em\relax PMLR, Jun. 2019, pp. 6105--6114. [Online]. Available: \url{https://proceedings.mlr.press/v97/tan19a.html}
\BIBentrySTDinterwordspacing

\bibitem{caesar2020nuscenes}
H.~Caesar, V.~Bankiti, A.~H. Lang, S.~Vora, V.~E. Liong, Q.~Xu, A.~Krishnan, Y.~Pan, G.~Baldan, and O.~Beijbom, ``nuscenes: A multimodal dataset for autonomous driving,'' in \emph{Proceedings of the IEEE/CVF conference on computer vision and pattern recognition}, 2020, pp. 11\,621--11\,631.

\bibitem{Argoverse2}
B.~Wilson, W.~Qi, T.~Agarwal, J.~Lambert, J.~Singh, S.~Khandelwal, B.~Pan, R.~Kumar, A.~Hartnett, J.~K. Pontes, D.~Ramanan, P.~Carr, and J.~Hays, ``Argoverse 2: Next generation datasets for self-driving perception and forecasting,'' in \emph{Proceedings of the Neural Information Processing Systems Track on Datasets and Benchmarks (NeurIPS Datasets and Benchmarks 2021)}, 2021.

\bibitem{he2016deep}
K.~He, X.~Zhang, S.~Ren, and J.~Sun, ``Deep residual learning for image recognition,'' in \emph{Proceedings of the IEEE conference on computer vision and pattern recognition}, 2016, pp. 770--778.

\bibitem{loshchilov2018decoupled}
I.~Loshchilov and F.~Hutter, ``Decoupled weight decay regularization,'' in \emph{International Conference on Learning Representations}, 2018.

\bibitem{chen2022gkt}
S.~Chen, T.~Cheng, X.~Wang, W.~Meng, Q.~Zhang, and W.~Liu, ``Efficient and robust 2d-to-bev representation learning via geometry-guided kernel transformer,'' \emph{arXiv preprint arXiv:2206.04584}, 2022.

\bibitem{yuan2024streammapnet}
T.~Yuan, Y.~Liu, Y.~Wang, Y.~Wang, and H.~Zhao, ``Streammapnet: Streaming mapping network for vectorized online hd map construction,'' in \emph{Proceedings of the IEEE/CVF Winter Conference on Applications of Computer Vision}, 2024, pp. 7356--7365.

\end{thebibliography}


% \end{IEEEbiography}


% % insert where needed to balance the two columns on the last page with
% %\newpage



%\vfill


\end{document}